\documentclass[letterpaper]{article} 
\usepackage{aaai25}  
\usepackage{times}  
\usepackage{helvet}  
\usepackage{courier}  
\usepackage[hyphens]{url}  
\usepackage{graphicx} 
\usepackage{natbib}  
\usepackage{caption} 
\usepackage{xcolor}   
\usepackage{paralist}
\usepackage{amsmath,amssymb,amsfonts}
\usepackage{algorithm}
\usepackage{comment}
\usepackage{yhmath}
\usepackage[noend]{algpseudocode}
\usepackage{enumitem}
\usepackage{subfig}
\usepackage{newfloat}
\usepackage{listings}
\usepackage[utf8]{inputenc}
\usepackage[english]{babel}
\usepackage[T1]{fontenc}
\usepackage{mathtools}
\DeclareCaptionStyle{ruled}{labelfont=normalfont,labelsep=colon,strut=off} 
\floatstyle{ruled}
\newfloat{listing}{tb}{lst}{}
\floatname{listing}{Listing}
\def\L#1{\raise .10ex\hbox{\scriptsize\tt #1}&}
\def\LHS#1{\hbox to .80em{#1\hfill}}

\def\K#1{\textbf{#1}}
\def\N{\\[.5ex]}

\def\I{\hspace{1.5em}}

\def\BB{{\mathcal B}} 
\def\vb{v_{\mathcal{B}}} 
\def\vp{v_\pi} 
\def\vi{v_\mathcal{I}} 
\def\mycirc\csname#1\endcsname{%
	\begin{tikzpicture}[baseline=(C.base)]
		\node[draw,circle,inner sep=1pt](C) {\csname#1\endcsname};
\end{tikzpicture}}
\newcommand*\mycirct[1]{%
	\begin{tikzpicture}[baseline=(C.base)]
		\node[draw,circle,inner sep=0.5pt](C) {#1};
\end{tikzpicture}}

\newtheorem{proposition}{Proposition}

\DeclareCaptionStyle{ruled}{labelfont=normalfont,labelsep=colon,strut=off} 
\floatstyle{ruled}
\newfloat{listing}{tb}{lst}{}
\floatname{listing}{Listing}
\title{Neural Control and Certificate Repair via Runtime Monitoring}
\author{
    Emily Yu\textsuperscript{\rm 1}, \DJ or\dj e $\mathbf{\check{\text{Z}}}$ikeli\'c\textsuperscript{\rm 2}, Thomas A. Henzinger\textsuperscript{\rm 1}
}
\affiliations{
    \textsuperscript{\rm 1}Institute of Science and Technology Austria, Klosterneuburg, Austria\\
    \textsuperscript{\rm 2}Singapore Management University, Singapore, Singapore
}

\usepackage{bibentry}

\begin{document}

\maketitle

\begin{abstract}
Learning-based methods provide a promising approach to solving highly non-linear control tasks that are often challenging for classical control methods. To ensure the satisfaction of a safety property, learning-based methods jointly learn a control policy together with a certificate function for the property. Popular examples include barrier functions for safety and Lyapunov functions for asymptotic stability. While there has been significant progress on learning-based control with certificate functions in the white-box setting, where the correctness of the certificate function can be formally verified, there has been little work on ensuring their reliability in the black-box setting where the system dynamics are unknown.
In this work, we consider the problems of certifying and repairing neural network control policies and certificate functions in the black-box setting. We propose a novel framework that utilizes runtime monitoring to detect system behaviors that violate the property of interest under some initially trained neural network policy and certificate. These violating behaviors are used to extract new training data, that is used to re-train the neural network policy and the certificate function and to ultimately repair them. We demonstrate the effectiveness of our approach empirically by using it to repair and to boost the safety rate of neural network policies learned by a state-of-the-art method for learning-based control on two autonomous system control tasks.
\end{abstract}
\section{Introduction}

The rapid advance of machine learning has sparked interest in using it to solve hard problems across various application domains, and autonomous robotic systems control is no exception. Learning-based control methods obtain data through repeated interaction with an unknown environment, which is then used to learn a control policy. A popular example is (deep) reinforcement learning algorithms, which learn neural network policies~\cite{MnihKSRVBGRFOPB15,sutton2018reinforcement}. However, the complex and often uninterpretable nature of learned policies poses a significant barrier to their safe and trustworthy deployment in safety-critical applications such as self-driving cars or healthcare devices~\cite{AmodeiOSCSM16,GarciaF15}.

In control theory, a classical method to certify the correctness of a control policy concerning some property of interest is to compute a certificate function for that property~\cite{DawsonGF23}. A {\em certificate function} is a mathematical object which proves that the system under the control policy indeed satisfies the property. Common examples of certificate functions include Lyapunov functions for stability~\cite{khalil2002control} and barrier functions for safety set invariance~\cite{PrajnaJ04}. While certificate functions are a classical and well-established concept in control theory, earlier methods for their computation are based either on polynomial optimization or rely on manually provided certificates, which makes them restricted to polynomial systems and typically scalable to low-dimensional environments~\cite{AhmadiM16,DawsonGF23,SrinivasanAN021}. 

Recent years have seen significant progress in overcoming these limitations, by utilizing machine learning to compute certificates for {\em non-polynomial environments} with {\em neural network control policies}~\cite{AbateAGP21,ChangRG19,DawsonQGF21,DBLP:journals/ral/QinSF22,DBLP:conf/iclr/QinZCCF21,RichardsB018}. The key idea behind these methods is to {\em jointly learn} a policy together with a certificate function, with both parametrized as neural networks. This is achieved by defining a loss function that encodes all defining properties of the certificate function. The learning algorithm then incurs loss whenever some of the defining certificate conditions are violated at any system state explored by the learning algorithm. Hence, the training process effectively guides the learning algorithm towards a control policy that admits a correct certificate function. A survey of the existing approaches and recent advances can be found in~\cite{DawsonGF23}.
\begin{figure}[t]\centering
	\scalebox{0.6}{
		\includegraphics[]{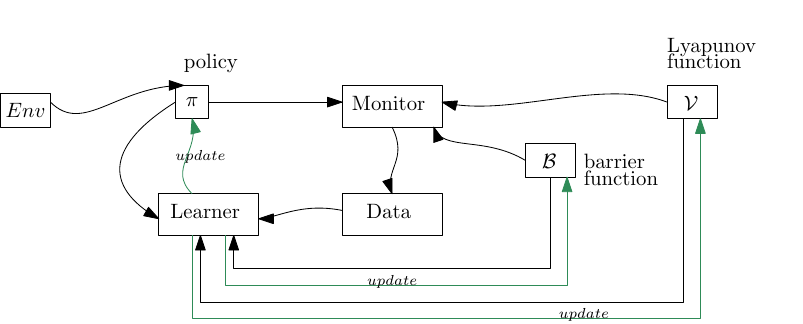}}
	\caption{The monitor-learner framework.\label{fig:framework}}
	\vspace{-1em}
\end{figure}

While the above approach {\em guides} the learning algorithm towards a correct certificate, it {does not guarantee} that the output control policy or the certificate function is correct. To address this issue, a number of recent works have considered the {\em white-box setting} which assumes complete knowledge of the environment dynamics and formally verifies the correctness of the certificate function. Notable examples include methods for learning and verifying neural Lyapunov functions~\cite{AbateAGP21,ChangRG19,RichardsB018} and neural barrier functions~\cite{AbateAEGP21,Zhang0V023,DBLP:conf/hybrid/ZhaoZC020}. More recently, to incorporate environment uncertainty, several works have gone a step beyond and considered neural certificates in possibly stochastic environments~\cite{AbateEGPR23,AnsaripourCHLZ23,ChatterjeeHLZ23,LechnerZCH22,MathiesenCL23,MazouzMRLL22,ZhiWLOZ24,ZikelicLHC23,ZikelicLVCH23}, assuming complete knowledge of the uncertainty model. 
While there are several methods that learn control policies together with certificates in the black-box setting, e.g.~\cite{DawsonQGF21,DBLP:journals/ral/QinSF22,DBLP:conf/iclr/QinZCCF21}, to the best of our knowledge no prior method considers the problem of analyzing and certifying their correctness in the {\em black-box setting}, without assuming any knowledge on the dynamics. In this setting, one cannot guarantee the correctness of certificate functions at every system state as we do not have access to the system model at each state. However, we may still use more lightweight techniques to evaluate the control policy and the certificate and to repair them in cases when they are found to be incorrect.

\smallskip\noindent{\bf Our contributions.} In this work, we propose a method for analyzing and repairing neural network policies and certificates in the {black-box setting}. While guaranteeing correctness in the black-box setting is not possible without further assumptions on the dynamics, our method is based on theoretical foundations of provably correct control as it uses both neural policies and certificates of their correctness in order to achieve successful repair.
More specifically, our method uses {\em runtime monitoring} to detect system behaviors that violate the property of interest or the certificate defining conditions. These behaviors are used to extract additional training data, which is then used by the training algorithm to re-train and ultimately {\em repair} the neural policy and the certificate. Our method consists of two modules -- called the {\em monitor} and the {\em learner}, which are composed in a loop as shown in Fig.~\ref{fig:framework}. The loop is executed until the monitor can no longer find property or certificate condition violating behaviors, upon which our algorithm concludes that the neural policy and the certificate have been repaired.

While the idea of using runtime monitors to identify incorrect behaviors is natural, it introduces several subtle challenges whose overcoming turns out to be highly non-trivial. The first and obvious approach to runtime monitoring of a control policy in isolation is to simply test it, flag runs that violate the property, and add visited states to the training set. However, such a simple monitor suffers from two significant limitations. First, it can only be used to monitor violations of properties that can be observed from finite trajectories such as safety, but not of properties that are defined by the limiting behavior such as stability. Second, the simple monitor detects property violations only {\em after} they happen, meaning that it cannot identify states and actions that do not yet violate the property but inevitably lead to property violation. In contrast, in practice we are often able to predict in advance when some unsafe scenario may occur. For instance, if we drive a car in the direction of a wall at a high speed, then even before hitting the wall we can predict that a safety violation would occur unless we change the course of action.

To overcome these limitations, we advocate the runtime monitoring of neural network policies {\em together} with certificate functions and propose two new monitoring algorithms. We are not aware of any prior method that utilizes certificate functions for the runtime monitoring of control policies. The first algorithm, called {\em Certificate Policy Monitor (CertPM)}, issues a warning whenever either the property or some of the defining conditions of the certificate are violated. The second algorithm, called {\em Predictive Policy Monitor (PredPM)}, goes a step further by estimating the remaining time until the property or some certificate condition may be violated, and issues a warning if this time is below some tolerable threshold. Experimental results demonstrate the ability of CertPM and PredPM to detect the property or certificate condition violating behaviors. Furthermore, we show that CertPM and PredPM allow our method to {\em repair} and significantly improve neural policies and certificates computed by a state-of-the-art learning-based control theory method.

Our contributions can be summarized as follows:
\begin{compactenum}
	\item {\em Runtime monitoring of policies and certificates.} We design two novel algorithms for the runtime monitoring of control policies and certificates in the {\em black-box setting}. 
	\item {\em Monitoring-based policy and certificate repair.} By extracting additional training data from warnings issued by our monitors, we design a novel method for {\em automated repair} of neural network control policies and certificates. 
	\item {\em Empirical evaluation.} Our prototype successfully repairs neural network control policies and certificates computed by a state-of-the-art learning-based control method.
\end{compactenum}


\smallskip\noindent{\bf Related work.}  Related works on learning-based control with certificate functions, as well as on formal verification of learned control policies and certificates in the white-box setting, have been discussed above, so we omit repetition. In what follows, we discuss some other related lines of work.

Constrained reinforcement learning (RL) algorithms consider constrained Markov decision processes (CMDPs)~\citep{altman1999constrained,Geibel06} and are also concerned with learning policies under safety constraints. An agent in a CMDP needs to maximize expected reward while satisfying hard constraints on expected cost for auxiliary cost functions. Notable examples of algorithms for solving CMDPs include Constrained Policy Optimization (CPO)~\citep{achiam2017constrained} or the method of~\cite{ChowNDG18} which uses a Lyapunov based approach.
While these algorithms perform well, they are empirical in nature and do not provide any mechanism for checking correctness of learned policies.

Shielding~\cite{DBLP:conf/tacas/BloemKKW15,AlshiekhBEKNT18} is an approach to ensuring the correctness of control policies with respect to safety properties that also utilizes runtime monitoring~\cite{DBLP:conf/isola/KonighoferL0B20,DBLP:conf/amcc/PrangerKTD0B21,DBLP:conf/aaai/Carr0JT23}. These methods use a monitor to detect if the system is close to reaching an unsafe set of states and thus violating the safety property. In such cases, the system switches to a {\em back-up policy} which is known to be safe and moves the system away from the unsafe region. A back-up policy can also be designed by reasoning about certificates, e.g.~\cite{ZhuXMJ19}. However, there are two fundamental differences between our method and shielding. First, our method does not assume the existence of a backup policy. Second, our method can also be used to {\em repair} a control policy, in cases when it is determined to be incorrect. Apart from shielding, runtime monitoring has been extensively used in the field of runtime verification as a more lightweight alternative to formal verification or when errors are only triggered at runtime~\cite{DBLP:conf/rv/FalconeP19,DBLP:journals/mscs/RenardFRJM19,DBLP:conf/rv/ZhouGKKL20}. Runtime monitoring and repair of neural network policies, but without using certificates, was considered in~\cite{BauerMarquartB22,LyuSZ23}.

\section{Preliminaries}\label{sec:prelims}

We consider a (deterministic, continuous-time) dynamical system $\Sigma = (X, U, f)$, where $\mathcal{X}\subseteq \mathbb{R}^n$ is the state space, $\mathcal{U}\subseteq \mathbb{R}^m$ is the control action space and $f: \mathcal{X}\times \mathcal{U}\rightarrow \mathcal{X}$ is the system dynamics, assumed to be Lipschitz continuous. The dynamics of the system are defined by
$\dot{x}(t)=f(x(t), u(t))$,
where $t \in \mathbb{R}_{\geq 0}$ is time, $x: \mathbb{R}_{\geq 0}\rightarrow \mathcal{X}$ is the state trajectory, and $u: \mathbb{R}_{\geq 0}\rightarrow \mathcal{U}$ is the control input trajectory. The control input trajectory is determined by a control policy $\pi: \mathcal{X} \rightarrow \mathcal{U}$. We use $ \mathcal{X}_0\subseteq  \mathcal{X}$ to denote the set of initial states of the system. For each initial state $x_0 \in \mathcal{X}_0$, the dynamical system under policy $\pi$ gives rise to a unique trajectory $x(t)$ with $x(0) = x_0$.  Control tasks are concerned with computing a control policy $\pi: \mathcal{X} \rightarrow \mathcal{U}$ such that, under the control input $u(t) = \pi(x(t))$, the state trajectory induced by $\dot{x}(t)=f(x(t), \pi(x(t))$ satisfies a desired property for every initial state in $\mathcal{X}_0$. In this work, we consider two of the most common families of control properties: 
\begin{compactenum}
	\item {\em Safety.} Given a set of unsafe states $\mathcal{X}_u \subseteq \mathcal{X}$, the dynamical system satisfies the {\em safety} property under policy~$\pi$ with respect to $\mathcal{X}_u$, if it never reaches $\mathcal{X}_u$, i.e.,~if for each initial state in $\mathcal{X}_0$ we have $x(t) \not\in \mathcal{X}_u$ for all $t \geq 0$.
	\item {\em Stability.} Given a set of goal states $\mathcal{X}_g \subseteq \mathcal{X}$, the dynamical system satisfies the {\em (asymptotic) stability} property under policy $\pi$ with respect to $\mathcal{X}_g$, if it asympotically converges to $\mathcal{X}_g$, i.e.,~if $\lim_{t\rightarrow\infty} \inf_{x_g \in \mathcal{X}_g} || x(t)-x_g ||=0$ for each initial state in $\mathcal{X}_0$.
\end{compactenum}
We also consider the stability-while-avoid property, which is a logical conjunction of stability and safety properties. To ensure that the dynamical system satisfies the property of interest, classical control methods compute a {\em certificate function} for the property. In this work, we consider barrier functions~\cite{DBLP:conf/cdc/AmesGT14} for proving safety and Lyapunov functions~\cite{khalil2002control} for proving stability. The stability-while-avoid property is then proved by computing both certificate functions.

\begin{proposition}[Barrier functions]\label{bs}
	Suppose that there exists a continuously differentiable function $\BB:  \mathcal{X}\rightarrow \mathbb{R}$ for the dynamical system $\Sigma$ under a policy $\pi$ with respect to the unsafe set $\mathcal{X}_u$, that satisfies the following conditions:
	\begin{compactenum}
		\item \label{bs:1} {\em Initial condition.} $\BB(x)\geq 0$ for all $x\in  \mathcal{X}_0$.
		\item {\em Safety condition.} $\BB(x)<0$ for all $x\in  \mathcal{X}_u$.
		\item \label{bs:nd} {\em Non-decreasing condition.} $L_f \BB(x) + \BB(x)\geq 0$ for all $x\in \{x\mid \BB(x)\geq 0\}$, where $L_f \BB = \frac{\partial \BB}{\partial x}f(x,u)$ is the Lie derivative of $\BB$ with respect to $f$.
	\end{compactenum}
	Then, $\Sigma$ satisfies the safety property under $\pi$ with respect to $\mathcal{X}_u$, and we call $\BB$ a {\em barrier function}.
\end{proposition}

\begin{proposition}[Lyapunov functions]\label{lc}
	Suppose that there exists a continuously differentiable function $\mathcal{V}:  \mathcal{X}\rightarrow \mathbb{R}$ for the dynamical system $\Sigma$ under policy $\pi$ with respect to the goal set $\mathcal{X}_g$, that satisfies the following conditions:
	\begin{compactenum}
		\item \label{lc:1} {\em Zero upon goal.} $\mathcal{V}(x_g)= 0$ for all $x_g \in \mathcal{X}_g$.
		\item {\em Strict positivity away from goal.} $\mathcal{V}(x)>0$ for all $x\in  \mathcal{X}\backslash\mathcal{X}_g$.
		\item \label{lc:3} {\em Decreasing condition.} $L_f\mathcal{V} < 0$ for all $x\in \mathcal{X}\backslash\mathcal{X}_g.$
	\end{compactenum}
	Then, $\Sigma$ satisfies the stability property under $\pi$ with respect to $\mathcal{X}_g$, and we call $\mathcal{V}$ a {\em Lyapunov function}. 
\end{proposition}


\noindent{\bf Assumptions.} We consider the black-box setting, meaning that the state space $\mathcal{X}$, the control action space $\mathcal{U}$ as well as the goal set $\mathcal{X}_g$ and/or the unsafe set $\mathcal{X}_u$ (depending on the property of interest) are {\em known}. However, the system dynamics function $f$ is {\em unknown} and we only assume access to an engine which allows us to execute the dynamics function.

\smallskip\noindent{\bf Problem statement.} Consider a dynamical system $\Sigma$ defined as above. Suppose we are given one of the above properties, specified by the unsafe set $\mathcal{X}_u$ and/or the goal set $\mathcal{X}_g$. Moreover, suppose that we are given a control policy $\pi$ and a certificate function $\mathcal{B}$ for the property.
\begin{compactenum}
	\item {\em Policy certification and repair.}  Determine whether the system $\Sigma$ under policy $\pi$ satisfies the property. If not, repair the policy $\pi$ such that the property is satisfied.
	\item {\em Certificate certification and repair.} Determine whether $\mathcal{B}$ is a good certificate function which shows that the dynamical system $\Sigma$ under policy $\pi$ satisfies the property. If not, repair the certificate function $\mathcal{B}$ such that it becomes a correct certificate function.
\end{compactenum}

\section{Runtime Monitoring Policies and Certificates}

We now present our algorithms for runtime monitoring of a policy by monitoring it together with a certificate function. These algorithms present the first step in our solution to the two problems defined above. The second step, namely policy and certificate repair, will follow in the next section. The runtime monitoring algorithms apply to general policies and certificate functions, not necessarily being neural networks. 

\smallskip\noindent{\bf Motivation for certificate monitoring.} Given a dynamical system $\Sigma = (\mathcal{X}, \mathcal{U}, f)$ and a property of interest, a {\em monitor} is a function $\mathcal{M}: \mathcal{X}^+ \rightarrow \{0,1\}$ that maps each finite sequence of system states to a verdict on whether the sequence violates the property. This means that the monitor can only detect violations with respect to properties whose violations can be observed from finite state trajectories. This includes safety properties where violations can be observed upon reaching the unsafe set, but {\em not} infinite-time horizon properties like stability which requires the state trajectory to asymptotically converge to the goal set {\em in the limit}. 
In contrast, monitoring both the control policy and the certificate function allows the monitor to issue a verdict on either (1)~property violation, or (2)~certificate violation, i.e.~violation of one of the defining conditions in Proposition~\ref{bs} for barrier functions or Proposition~\ref{lc} for Lyapunov functions. By Propositions~\ref{bs} and~\ref{lc}, we know that the barrier function or the Lyapunov function being correct provides a formal proof of the safety or the stability property. Hence, in order to show that there are no property violations and that the property of interest is satisfied, it suffices to show that there are no certificate violations which can be achieved by monitoring both the control policy and the certificate function.

\subsection{Certificate Policy Monitor}

We call our first monitor the {\em certificate policy monitor (CertPM)}. For each finite sequence of observed states $x_0, x_1, \dots, x_n$, the monitor $\mathcal{M}_{\textsf{CertPM}}$ issues a verdict on whether a property or a certificate violation has been observed. If we are considering a safety property, the property violation verdict is issued whenever $x_n \in \mathcal{X}_u$, and the certificate violation verdict is issued whenever one of the three defining conditions in Proposition~\ref{bs} is violated at $x_n$. If we are considering a stability property, the property violation verdict cannot be issued, however, the certificate violation verdict is issued whenever one of the defining conditions in Proposition~\ref{lc} is violated at $x_n$. 
In the interest of space, in what follows we define the monitor $\mathcal{M}_{\textsf{CertPM}}$ for a safety property specified by the unsafe set of states $\mathcal{X}_u$. The definition of $\mathcal{M}_{\textsf{CertPM}}$ for a stability property is similar, see the Appendix. Let $\pi$ be a policy and $\BB$ be a barrier function: 
\begin{compactitem}
	\item The safety violation verdict is issued if $x_n \in \mathcal{X}_u$. We set $\mathcal{M}_{\textsf{CertPM}}(x_0,x_1,\dots,x_n) = 1$.
	\item The certificate violation verdict for the Initial condition in Proposition~\ref{bs} is issued if $x_n \in \mathcal{X}_0$ is an initial state but $\BB(x_n) < 0$. We set $\mathcal{M}_{\textsf{CertPM}}(x_0,x_1,\dots,x_n) = 1$.
	\item The certificate violation verdict for the Safety condition in Proposition~\ref{bs} is issued if $\BB(x_n)<0$. We set $\mathcal{M}_{\textsf{CertPM}}(x_0,x_1,\dots,x_n) = 1$.
	\item Checking the Non-decreasing condition in Proposition~\ref{bs} is more challenging since the dynamical system evolves over continuous-time and the non-decreasing condition involves a Lie derivative. To address this challenge, we approximate the non-decreasing condition by considering the subsequent observed state $x_{n+1}$ and approximating the Lie derivative via $$\widehat{L_f\BB}(x_n)=\frac{\BB(x_{n+1}) -\BB(x_n)}{t_{n+1} - t_n}.$$
	This requires the monitor to wait for at least one new observation before issuing the verdict. The approximation satisfies $ |  \widehat{L_f\BB}(x) - L_f\BB(x)|\leq \epsilon$ for all $x\in \mathcal{X}$
	where $\epsilon=\frac{1}{2}\Delta_t(\mathcal{C}_\BB\mathcal{L}_f+\mathcal{C}_f\mathcal{L}_\BB)\mathcal{C}_f$, with $\mathcal{L}_\BB$ and $\mathcal{L}_f$ being the Lipschitz constants of $\BB$ and $f$ bounded by constants $\mathcal{C}_\BB, \mathcal{C}_f\in\mathbb{R}_{>0}$~\cite{DBLP:journals/tac/NejatiLJSZ23}. This suggests we can achieve good precision by using sufficiently small time intervals for monitoring. The certificate violation verdict for the Non-decreasing condition in Proposition~\ref{bs} is issued if $\BB(x_n) \geq 0$ but $\widehat{L_f\BB}(x_n) + \BB(x_n) < 0$. We set $\mathcal{M}_{\textsf{CertPM}}(x_0,x_1,\dots,x_n) = 1$.
	\item Otherwise, no violation verdict is issued. We set $\mathcal{M}_{\textsf{CertPM}}(x_0,x_1,\dots,x_n) = 0$.
\end{compactitem}

\subsection{Predictive Policy Monitor}

CertPM checks if the property or one of the certificate defining conditions is violated at the states observed so far. Our second monitor, which we call the {\em predictive policy monitor (PredPM)}, considers the case of safety properties and {\em estimates the remaining time} before the property or the certificate violation may occur in the future. PredPM then issues a verdict based on whether any of the estimated remaining times are below some pre-defined thresholds. Thus, our second monitor aims to predict future behavior and issue property and certificate violation verdicts {\em before they happen}.

Since PredPM is restricted to safety properties, let $\mathcal{X}_u$ be the set of unsafe states. Let $\pi$ be a policy and $\BB$ be a barrier function. For each finite sequence of observed states $x_0, x_1, \dots, x_n$, PredPM computes three quantitative assessments $[v_U, v_S, v_N]$, where:
\begin{compactenum}
	\item $v_U$ is an estimate of the time until $\mathcal{X}_u$ is reached and hence the safety property is violated (or, if $x_n\in\mathcal{X}_u$ already, $v_U$ is an estimate of the time until the safe part $\mathcal{X}\backslash \mathcal{X}_u$ is reached).
	\item $v_S$ is an estimate of the time until the set $\{x \in \mathcal{X} \mid \BB(x) < 0\}$ may be reached and hence the Safety condition in Proposition~\ref{bs} may be violated;
	\item $v_N$ is an estimate of the time until the Non-dec.~condition in Proposition~\ref{bs} may be violated.
\end{compactenum}
Note that these three values are estimates on the remaining time until some set $S \subseteq \mathcal{X}$ is reached, where $S = \mathcal{X}_u$ (or $S = \mathcal{X}\backslash\mathcal{X}_u$ if $x_n\in \mathcal{X}_u$) for computing $v_U$, $S = \{x \in \mathcal{X} \mid \BB(x) < 0\}$ for computing $v_S$, and $S = \{x \in \mathcal{X} \mid \widehat{L_f\BB}(x) + \BB(x) < 0\}$ for computing $v_N$. PredPM computes each of the three values by solving the following problem:
\begin{equation}\nonumber
	\begin{aligned}
		\min_a \; \mathcal{T} \;
		\textrm{s.t.}  \; \frac{dx}{dt} = v(t), \; \frac{dv}{dt} = a(t), \; |a(t)| \leq a_{max}, \\\forall t \in [0,\mathcal{T}];
		\; x(\mathcal{T}) \in S, \; x(0) = x_n, \; v(0) = v_0.
	\end{aligned}
\end{equation}
Here, $v(t)$ is the velocity and $a(t)$ is the acceleration of the trajectory $x(t)$, with $a_{max}$ being the maximum allowed acceleration. Intuitively, solving this optimization problem results in the acceleration that the controller should use at each time such that the set $S$ is reached in the shortest time possible. To make the computation physically more realistic, we assume a physical bound 
$a_{max}$ on the maximum acceleration that the controller can achieve at any time. To compute an approximate solution to this problem and hence estimate $v_U$, $v_S$ and $v_N$, PredPM discretizes the time $[0,\mathcal{T}]$ and for each discrete time point it uses stochastic gradient descent to select the next acceleration within the range $[0,a_{max}]$.


Once PredPM computes $v_U, v_S, v_N$, it checks if any of the values exceed the thresholds $\xi_U,\xi_S, \xi_N$ that are assumed to be provided by the user. The monitor issues the verdict $\mathcal{M}_{\textsf{PredPM}}(x_0,x_1,\dots,x_n) = 1$ if either $v_U > \xi_U$ or $v_S > \xi_S$ or $v_N > \xi_N$. Otherwise, it issues the verdict $0$. Having quantitative assessments allows the user to specify how fault-tolerant the monitor should be. For highly safety-critical systems, one can set the thresholds to be positive such that the monitor signals a warning {\em in advance}, i.e.,~when it predicts a dangerous situation and before it happens. 

\section{Neural Policy and Certificate Repair}\label{sec:repair}

\begin{figure}[t]
	\begin{tabular}{@{}r@{\hspace{.75em}}l@{}}
		\L{}\hspace{-1.25em}\textsc{PolicyRepair} (policy $\pi$, barrier function $\BB$, number $D$, \N
		\L{}\hspace{5em} time points $t_0 = 0 < t_1 < \dots < t_N$,\N
		\L{}\hspace{5em}  (optional) thresholds $\xi_U$, $\xi_S$ $\xi_N$ for PredPM)\N
		\\[-2.0ex]
		\L{1} $\mathcal{M}\leftarrow \textsc{buildMonitor}(\mathcal{X})$\N
		\L{} \footnotesize\textcolor{gray}{$\triangleright$ initializing a monitor, either CertPM or PredPM} \N
		\L{2} $D_\textrm{New-data}\leftarrow \emptyset$ \footnotesize\textcolor{gray}{$\triangleright$ new training data collected by the monitor}\N
		\L{3} $\tilde{\mathcal{X}}_0\leftarrow D$ states randomly sampled from $\mathcal{X}_0$\N 
		\L{4}\K{for} $x_0 \in \tilde{\mathcal{X}}_0$ \K{do}\N
		\L{5}\I $x(t)\leftarrow$ state trajectory from $x(0) = x_0$\N
		\L{6}\I\K{for} $n \in \{0,1,\dots,N\}$ \K{do}\N
		\L{7}\I\I $x_n\leftarrow$ observed state at time point $t_n$\N
		\L{8}\I\I \K{if} $\mathcal{M}(x_0,x_1,\dots,x_n) = 1$ \K{then} \N
		\L{9} \I\I\I$D_\textrm{New-data}\leftarrow D_\textrm{New-data} \cup \{x_n\}$\N
		\L{10} $D_{\textrm{Init}}^{\textrm{repair}}, D_{\textrm{Safe}}^{\textrm{repair}}, D_{\textrm{Non-dec}}^{\textrm{repair}} \leftarrow D_{\textrm{New-data}} \cap \mathcal{X}_0,$\N \L{}\I\I\I$D_{\textrm{New-data}} \cap \mathcal{X}_u, D_{\textrm{New-data}} \cap \{x \mid \BB(x) \geq 0\}$\N
		\L{11}\K{if} policy certification and repair \K{then} \footnotesize\textcolor{gray}{$\triangleright$ Problem~1}\N
		\L{12}\I $\pi, \BB \leftarrow$ repair with loss function in eq.~\eqref{eq:loss} \N
		\L{}\I\I\I and training data $D_{\textrm{Init}}^{\textrm{repair}}$, $D_{\textrm{Safe}}^{\textrm{repair}}$, $D_{\textrm{Non-dec}}^{\textrm{repair}}$ \N
		\L{13}\K{else if} certificate certification and repair \K{then} \footnotesize\textcolor{gray}{$\triangleright$ Problem~2}\N
		\L{14}\I $\BB \leftarrow$ repair with loss function in eq.~\eqref{eq:loss}\N
		\L{}\I\I and training data $D_{\textrm{Init}}^{\textrm{repair}}$,  $D_{\textrm{Safe}}^{\textrm{repair}}$, $D_{\textrm{Non-dec}}^{\textrm{repair}}$
	\end{tabular}
	\captionof{algorithm}{Monitoring-based neural network policy and certificate repair for safety properties.}
	\label{alg:monitorrepair}\vspace{-1em}
\end{figure}

Our monitors in the previous section are able to flag finite state trajectories that may lead to property or certificate violations. We now show how the verdicts issued by our monitors can be used to extract additional training data that describe the property or certificate violations. Hence, if the control policy and the certificate are both learned as neural networks, we show how they can be retrained on this new data and {\em repaired}. In the interest of space, in this section, we consider the case of safety properties where certificates are barrier functions. Suppose that $\mathcal{X}_u$ is a set of unsafe states in a dynamical system $\Sigma = (\mathcal{X}, \mathcal{U}, f)$. The pseudocode of our monitoring-based repair algorithm is shown in Algorithm~\ref{alg:monitorrepair}. Our method can be easily modified to allow Lyapunov function repair and we provide this extension in the Appendix.

\smallskip\noindent{\bf Learning-based control with certificates.} Before showing how to extract new training data and use it for policy and certificate repair, we first provide an overview of the general framework for jointly learning neural network policies and barrier functions employed by existing learning-based control methods~\cite{DBLP:conf/hybrid/ZhaoZC020}. Two neural networks are learned simultaneously, by minimizing a loss function which captures each of the defining conditions of barrier functions in Proposition~\ref{bs}. That way, the learning process is guided towards learning a neural control policy that admits a barrier function and hence satisfies the safety property. The loss function contains one loss term for each defining condition:
\begin{equation}\label{eq:loss}
	\mathcal{L}(\theta, \nu)=\mathcal{L}_{\textrm{Init}}(\theta, \nu) +\mathcal{L}_{\textrm{Safe}}(\theta,\nu)+\mathcal{L}_{\textrm{Non-dec}}(\theta, \nu),
\end{equation}
where $\theta$ and $\nu$ are vectors of parameters of neural networks $\pi_\theta$ and $\BB_\nu$, respectively, and
\begin{equation}\nonumber
	\begin{aligned}
		\mathcal{L}_{\textrm{Init}}(\theta,\nu) &= \frac{1}{|D_{\textrm{Init}}|} \sum_{x\in D_{\textrm{Init}}}\max(-\mathcal{B}_\nu(x),0);\\
		\mathcal{L}_{\textrm{Safe}}(\theta,\nu) &= \frac{1}{|D_{\textrm{Safe}}|} \sum_{x\in D_{\textrm{Safe}}}\max(\mathcal{B}_\nu(x),0);
	\end{aligned}
\end{equation}
\begin{equation}\nonumber
	\begin{aligned}
		\mathcal{L}_{\textrm{Non-dec}}(\theta, \nu) &= \frac{1}{|D_{\textrm{Non-dec}}|}\cdot\\
		& \sum_{x\in D_{\textrm{Non-dec}}}\max\Big(-\widehat{L_{f_\theta}\mathcal{B}}_\nu(x)-\mathcal{B}_\nu(x),0\Big).
	\end{aligned}
\end{equation}
In words, $D_{\textrm{Init}}, D_{\textrm{Safe}}, D_{\textrm{Non-dec}} \subseteq \mathcal{X}$ are the training sets of system states used for each loss term that incurs loss whenever the defining condition in Proposition~\ref{bs} is violated. 
The term $\widehat{L_{f_\theta}\mathcal{B}}_\nu(x)$ in $\mathcal{L}_{\textrm{Non-dec}}$ is approximated by executing the system dynamics from state $x$ for a small time interval.
Since we are interested in repair and not the initialization of $\pi_\theta$ and $\BB_\nu$, we omit the details on how this training data is collected and refer the reader to~\cite{DBLP:journals/ral/QinSF22}.

\smallskip\noindent{\bf Monitoring-based neural policy and certificate repair.} We now show how the verdicts of our monitors are used to obtain new training data that describes the property or certificate violations, and how this new training data is used for the policy and certificate repair. Algorithm~\ref{alg:monitorrepair} takes as input the neural network control policy $\pi$ and neural network barrier function $\BB$, trained as above. It also takes as input a finite set of time points $t_0 = 0 < t_1 < \dots < t_N$ at which the monitor observes new states, as well as the number $D$ of state trajectories that it monitors. In addition, if the PredPM monitor is to be used, the algorithm also takes as input the three thresholds $\xi_U$, $\xi_S$, and $\xi_N$.

Algorithm~\ref{alg:monitorrepair} first initializes the monitor $\mathcal{M}$ by constructing either the CertPM or the PredPM (line~1), the set of new training data $D_{\textrm{New-data}}$ which is initially empty (line~2) and a set $\tilde{\mathcal{X}}_0 \subseteq \mathcal{X}_0$ of $D$ initial states obtained via sampling from the initial set $\mathcal{X}_0$ (line~3). Then, for each initial state $x_0 \in \tilde{\mathcal{X}}_0$, it executes the dynamical system to obtain a state trajectory $x(t)$ from $x(0) = x_0$ (line~5). For each time point $t_n$, a new state $x_n$ is observed (line~7) and the monitor verdict $\mathcal{M}(x_0,x_1,\dots,x_n)$ is computed (line~8). If $\mathcal{M}(x_0,x_1,\dots,x_n)=1$, i.e. if the monitor issues a property or certificate violation verdict, the new state $x_n$ is added to the new training dataset $D_{\textrm{New-data}}$ (line~8). Once the new data is collected, it is used to initialize the new training datasets $D_{\textrm{Init}}^{\textrm{repair}} = D_{\textrm{New-data}} \cap \mathcal{X}_0$, $D_{\textrm{Safe}}^{\textrm{repair}} = D_{\textrm{New-data}} \cap \mathcal{X}_u$ and $D_{\textrm{Non-dec}}^{\textrm{repair}} = D_{\textrm{New-data}} \cap \{x \mid \BB(x) \geq 0\}$ (line~9). Finally, the neural network policy $\pi$ and the neural network barrier function $\BB$ are repaired by being retrained on the loss function in eq.~\eqref{eq:loss} with the new training datasets (line~11). If we are only interested in repairing the barrier function $\BB$ for a fixed control policy $\pi$, only the network $\BB$ is repaired while keeping the parameters of $\pi$ fixed (line~13).

\section{Experimental Evaluation}\label{sec:experiments}

We implemented a prototype of our method in Python~3.6 as an extension of the SABLAS codebase~\cite{DBLP:journals/ral/QinSF22}. SABLAS is a state-of-the-art method and tool for learning-based control of neural network control policies with certificate functions. In order to evaluate our method, we consider the benchmarks from the SABLAS codebase, together with policies and certificate functions learned by the SABLAS method for these benchmarks. We then apply our method to these control policies and certificate functions in order to experimentally evaluate the ability of our method to detect incorrect behaviors and to repair them. Our goal is to answer the following three research questions (RQs):
\begin{compactitem}
	\item {\bf RQ1:} Is our method able to detect violating behavior and repair neural network policies and certificate functions? This RQ pertains to Problem 1.
	\item {\bf RQ2:} Is our method able to detect incorrect behavior and repair neural network certificate functions? This RQ pertains to Problem 2 in the Preliminaries Section.
	\item {\bf RQ3:} How strong is predictive power of PredPM monitor, that is, can it predict safety violations {\em ahead of time}?
\end{compactitem} 
\begin{table}[t]
	\centering
	\caption{Results on repairing the control policy and the barrier function for DroneEnv. For PredPM, we evaluate it with three threshold configurations [$\xi_U,\xi_S,\xi_N$] (we chose three configurations for which we observed different numbers $D_\textsc{New}$ of property or certificate violations; see the Appendix for results on more thresholds configurations). Column $D_\textsc{New}$ shows the total number of property or certificate violations found by each method. The safety rate (SR) is the proportion of time during which the agent stays away from the unsafe set calculated by $\frac{1}{T}\int^T_0 [x(t)\notin\mathcal{X}_u]\; dt$, and BR is the proportion of time during which the system is within the invariant set $\{x \mid \BB(x) \geq 0\}$, and NDR is the proportion of time during which the barrier function satisfies the non-decreasing condition. The best results for each column are in bold. All results are averaged over 50~further executions. 
		\label{tb}}
	\label{tb:policycomp}\scalebox{0.96}{
		\begin{tabular}{l|c|c|c|c}
			\hline
			& \#$D_\textsc{New}$& \textbf{SR} (\%) & \textbf{BR}  (\%)& \textbf{NDR}  (\%) \\
			\hline
			Initialized  &-& 93.99& 87.03& 45.38\\  
			\hline
			Baseline &878& 96.61& -& -\\ 
			\hline
			CertPM &548&\textbf{99.13}&\textbf{100.00} & {90.66}\\ 
			
			\hline
			PredPM \scriptsize{[0,0,-1]}&146&99.06& \textbf{100.00} &90.11\\	
			\hline
			PredPM
			\scriptsize{[0,1,-5]}&355&98.67& \textbf{100.00} &90.12\\
			\hline
			PredPM
			\scriptsize{[2,2,0]} &\textbf{1000}&{99.09} & \textbf{100.00} & \textbf{91.67}\\
			
			\hline
	\end{tabular}}	
\end{table}
\begin{figure}[t]\captionsetup[subfigure]{labelformat=empty}	\vspace{-0.3em}
	\subfloat[]{\includegraphics[width = 0.43\textwidth]{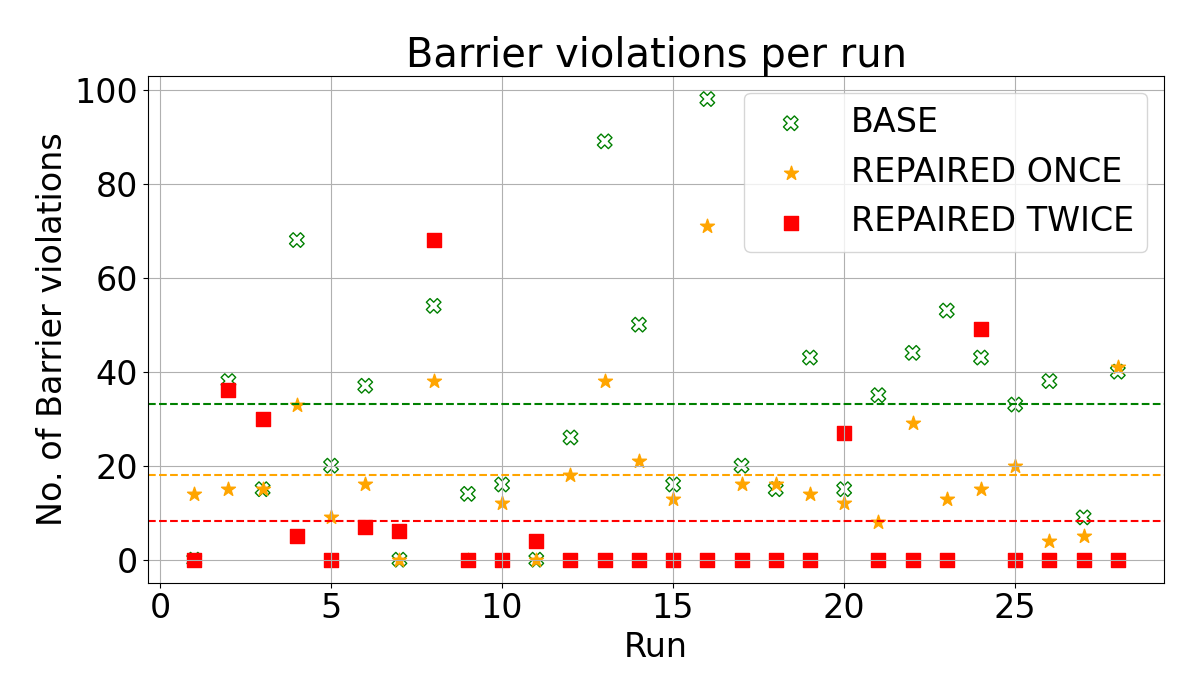}}\vspace{-3em} \\
	
	\subfloat[]{\includegraphics[width = 0.43\textwidth]{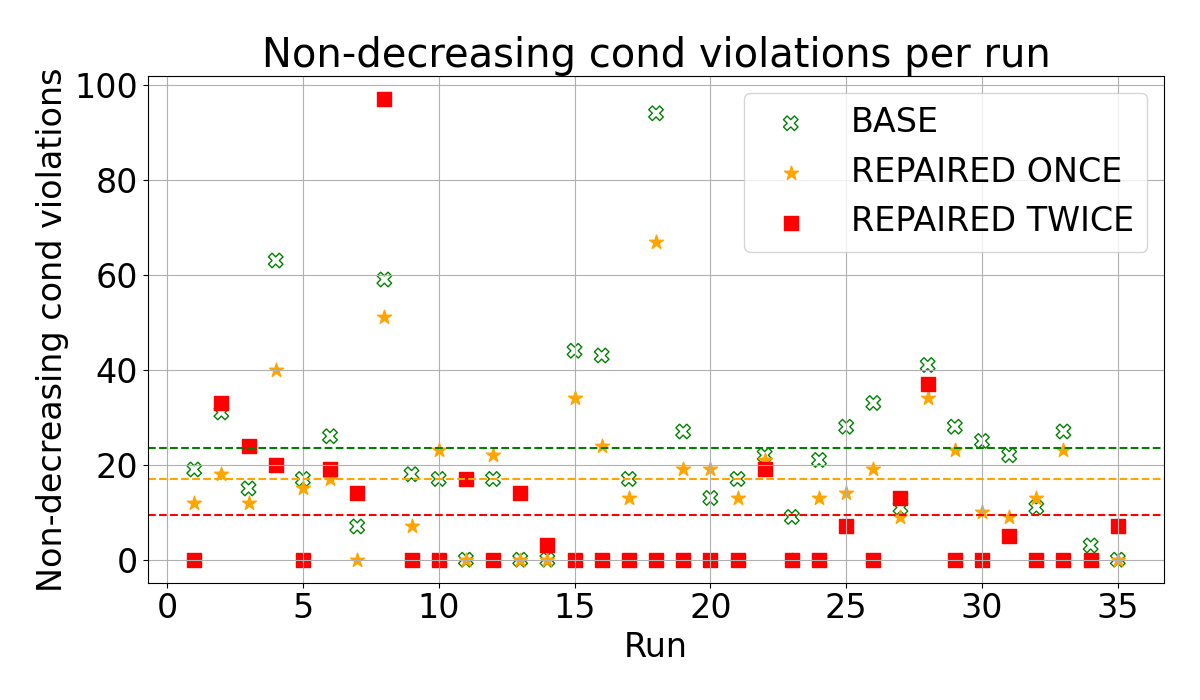}} 
	\vspace{-2em}
	\caption{The change in the number of certificate violations for the Safety condition and the Non-decreasing conditions of barrier functions, for the ship benchmark after one round of repair and after a second round of repair. The first round monitors $D = 15000$ system executions. The second round monitors an additional $D=20000$ executions. For better readability, we plot only the executions (out of 50) for which at least one certificate violation is detected.  \label{fig:certship}}
\end{figure}

\smallskip\noindent{\bf Benchmarks.} We consider two benchmarks that are available in the SABLAS codebase~\cite{DBLP:journals/ral/QinSF22}, originally taken from~\cite{fossen2000survey,DBLP:conf/iclr/QinZCCF21}. The first benchmark concerns a parcel delivery drone flying in a city (called the active drone), among 1024 other drones that are obstacles to be avoided. Only the active drone is controlled by the learned policy, whereas other drones move according to pre-defined routes. The state space of this environment is 8-dimensional and states are defined via $8$ variables $x=[x,y,z,v_x,v_y,v_z,\theta_x,\theta_y]$: three coordinate variables in the 3D space, three velocity variables, and two variables for row and pitch angles. The actions produced by the policy correspond to angular accelerations of $\theta_x,\theta_y$, as well as the vertical thrust. The drone is completing a delivery task at a set destination, hence the property of interest in this task is a stability-while-avoid property. 

The second benchmark ShipEnv concerns a ship moving in a river among 32 other ships. The state space of this environment is 6-dimensional and states are defined via $6$ variables $[x,y,\theta, u,v,w]$. The first two variables specify the 2D coordinates of the ship, $\theta$ is the heading angle, $u,v$ are the velocities in each direction, and $w$ is the angular velocity of the heading angle. The property of interest in this task is also a stability-while-avoid property with obstacles being collisions with other ships. 

\smallskip\noindent{\bf Experimental setup.} We consider the black-box setting, hence the dynamics are unknown to the monitor and repair algorithm. For each monitored execution, $N=2000$ and $N=1200$ states are observed for ShipEnv and DroneEnv, respectively, spaced out at time intervals of $\Delta_t=0.1s$. In each environment, the states of the eight nearest obstacles (i.e.~drones or ships) are given as observations to the policy, as well as the certificate functions. In our implementation of the PredPM monitor, we employ Adam~\cite{kingma2014adam} for approximating the assessments $[v_U, v_S, v_N]$.

\begin{figure}[t]\captionsetup[subfigure]{labelformat=empty}
	\centering
	\vspace{-1.4em}
	\subfloat[]{\includegraphics[width = 0.24\textwidth]{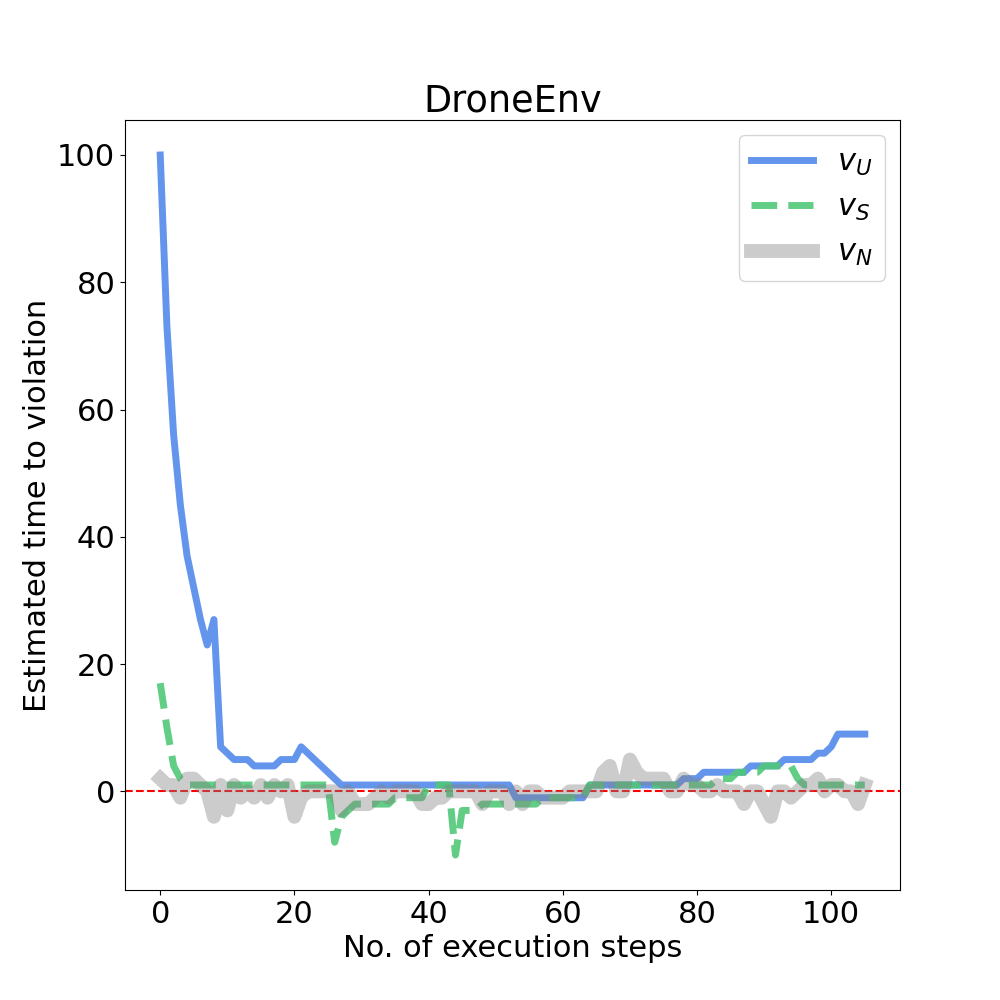}}
	\subfloat[]{\includegraphics[width = 0.24\textwidth]{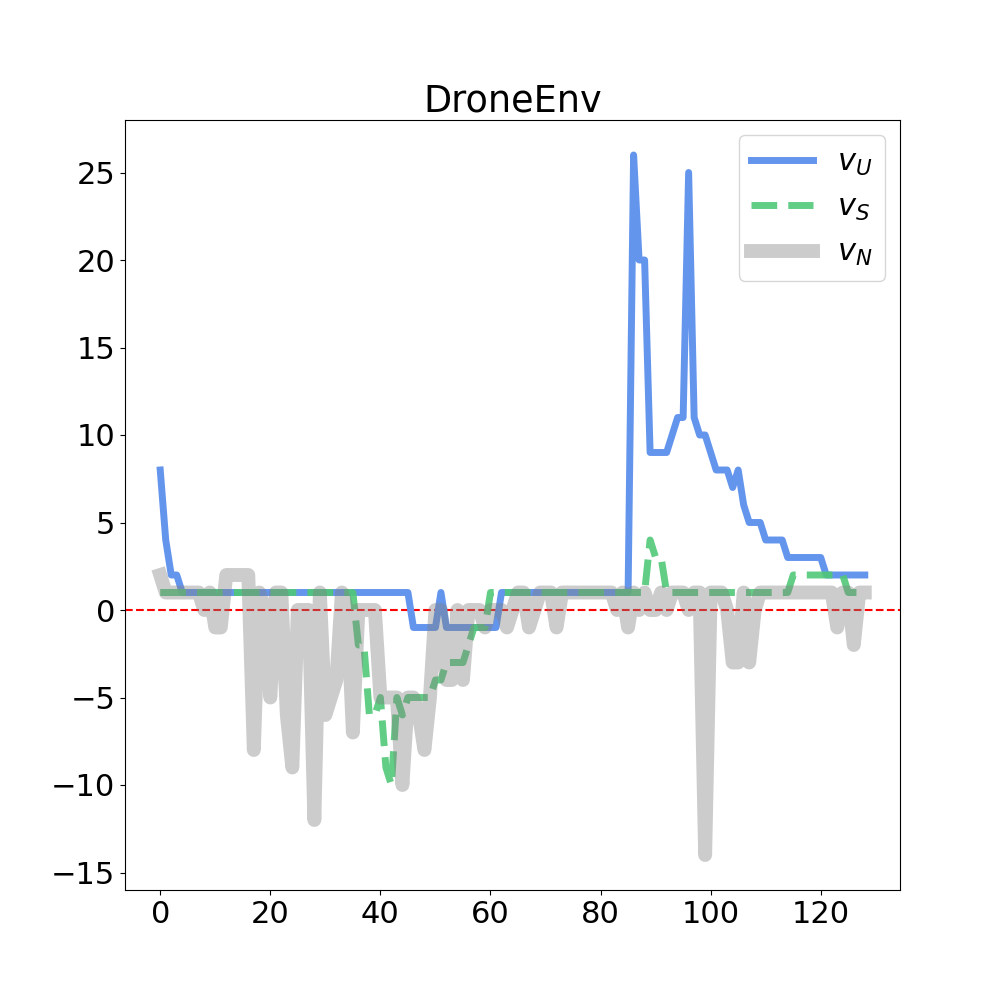}}
	\vspace{-1cm}
	\\
	\subfloat[]{\includegraphics[width = 0.24\textwidth]{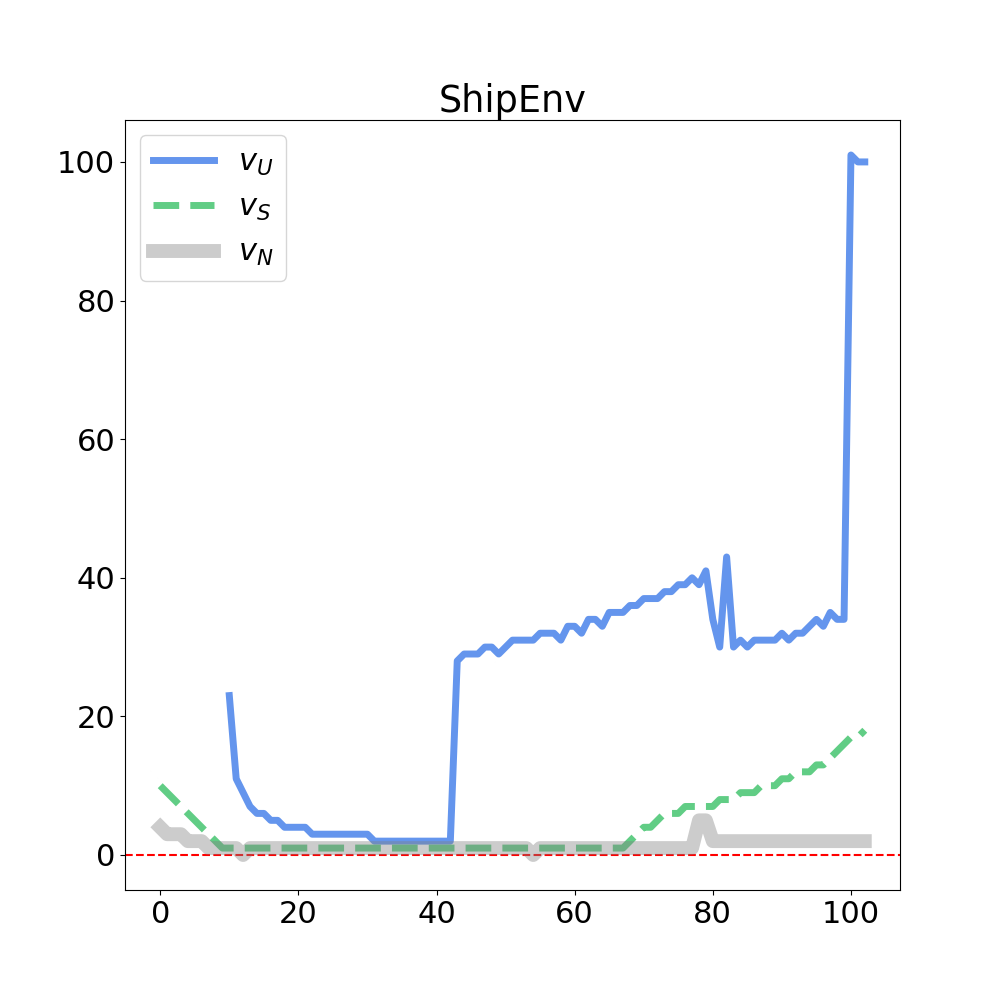}}
	\subfloat[]{\includegraphics[width = 0.24\textwidth]{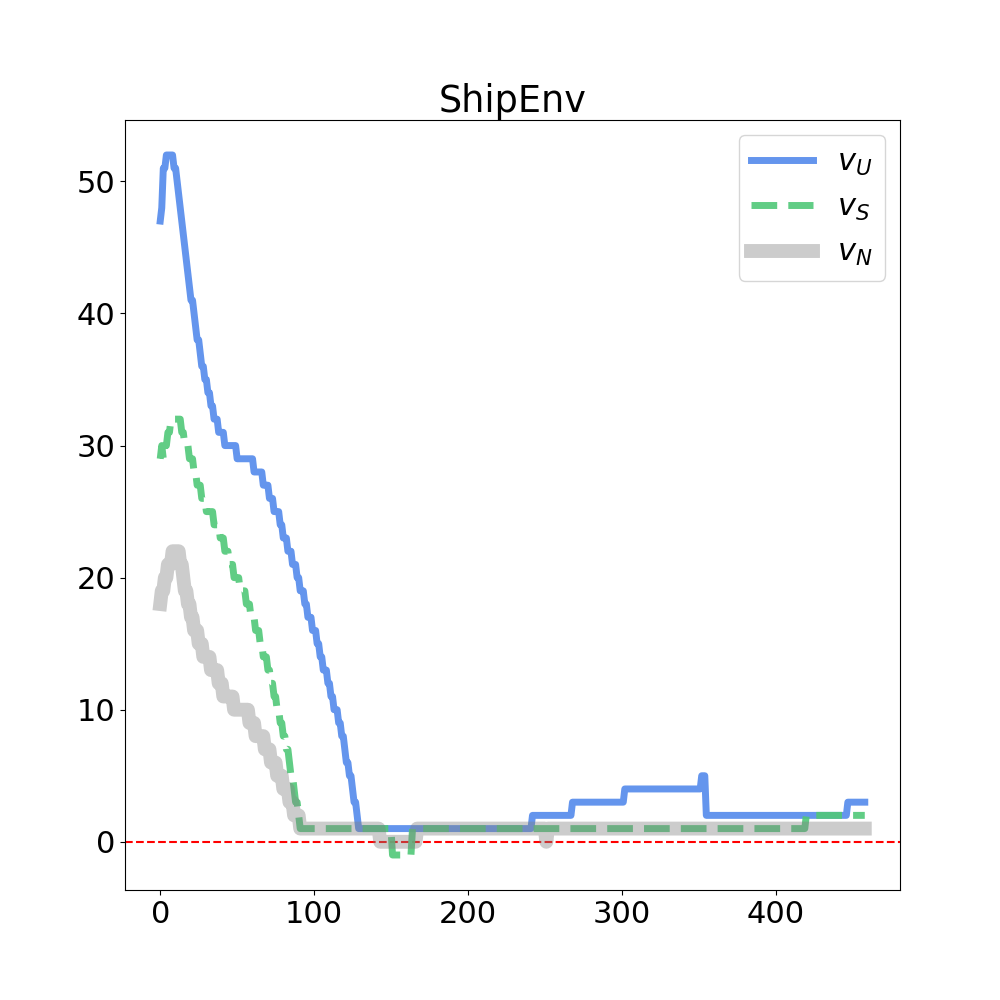}}
	\vspace{-2.2em}
	\caption{Estimates $v_U$, $v_S$, and $v_N$ for two different systems executions computed by the PredPM for the drone and the ship benchmarks. 
		\label{fig:rv-drone}}
\end{figure}

\smallskip\noindent{\bf Results: RQ1.} We observed that the control policy learned by SABLAS for DroneEnv does not satisfy the stability-while-avoid property on all runs -- its safety rate is $93.99\%$ whereas it leads to collision with other drones $6.01\%$ of the time, initialized with $10^4$ state samples. We applied our repair method to the control policy and the barrier function learned by SABLAS with $D=1000$. 
To evaluate the importance of monitoring both neural policies and certificates,
we also compare our method against the \textit{baseline} approach. The baseline is the simple monitor described in the Introduction, which only monitors a neural policy, flags traces that reach an unsafe state and adds these states to re-training data.
Our results are summarized in Table~\ref{tb}. As we can see, both CertPM and PredPM monitors are able to effectively repair the control policy and the barrier function and lead to significantly higher safety rates (SR), reaching $99.13\%$. 
The proportions of time at which the Safety and the Non-decreasing conditions of barrier functions are satisfied (BR and NDR in Table~\ref{tb}) go up from $87.03\%$ to $100.00\%$, and from $45.38\%$ to $91.67\%$. This demonstrates the advantage of monitoring both the control policy and the certificate toward effective and successful repair. Finally, the comparison between CertPM and PredPM shows that CertPM is slightly better in repairing the control policy, but PredPM is more effective for repairing the barrier function. Due to its predictive nature, PredPM identifies a larger number of property and certificate violations with the right choice of thresholds, resulting in larger re-training data $D_\textsc{New-data}$. We conducted an analogous experiment with a certificate function consisting both of a barrier and a Lyapunov function and again observed significant improvements upon repair: see the Appendix. The Appendix also provides results on more thresholds configurations of PredPM, beyond the three configurations considered in Table~\ref{tb}.


\smallskip\noindent{\bf Results: RQ2.} We observed that the policy learned by SABLAS for the ship benchmark already achieves SR close to $100.00\%$, however, the learned barrier function provides significantly lower BR and NDR. Hence, we use the ship benchmark to answer RQ2 which is concerned with the repair of a certificate function for a \emph{given} control policy. In this case, a good certificate function acts as a proof of correctness that allows more trustworthy policy deployment. Figure~\ref{fig:certship} shows what the number of certificate violations for the barrier function looks like before and after repair, for the ship benchmark with CertPM used as a monitor. The results demonstrate that there are significantly fewer certificate violations upon repair. Additionally, we conducted the same experiment for the drone benchmark, and also observed significant level of improvement in BR and NDR upon repair. We refer to the Appendix for more results.



\smallskip\noindent{\bf Results: RQ3.} Recall that PredPM considers safety properties and monitors the control policy together with the barrier function. Upon each new observation, it computes an estimate $v_U$ on the remaining time before the safety property may be violated, $v_S$ on the remaining time before the Safety condition of barrier functions may be violated, and $v_N$ on the remaining time before the Non-decreasing condition of barrier functions may be violated. Figure~\ref{fig:rv-drone} shows how these estimates change along two different executions for DroneEnv and ShipEnv. It can be seen that $v_S$ becomes negative before $v_U$, meaning the system is estimated to violate the Safety condition of barrier functions before it reaches the unsafe region. Hence, by tracking $v_S$, PredPM can predict unsafe behaviors and raise warnings \emph{before} they happen. In comparison to the others, there are more estimated violations of the Non-decreasing condition. This means that $\xi_N$ can be set to a lower value, as we tend to consider the other two violations to be more severe. Overall, the experiments suggest PredPM can be particularly well-suited for runtime use, in addition to repairing neural networks.

\smallskip\noindent{\bf Summary of results.} Our experimental results empirically justify the following claims: (i)~Our method is able to successfully repair neural network control policies and certificate functions. Using either CertPM or PredPM for repair leads to significant improvements over the initial policy; (ii)~Our method is able to successfully repair neural network certificate functions in the setting where the control policies are fixed; (iii)~Using PredPM allows predicting safety property violations {\em before they happen}, hence showing potential for practical safety deployment even in the runtime setting.

\smallskip\noindent{\bf Practical considerations and limitations.} To conclude, we also discuss two practical aspects that one should take into account before the deployment of our method:
(i) As highlighted in the Introduction, our method provides no guarantees on the correctness of repaired policies. This means that, in principle, one could end up with a policy whose performance is suboptimal compared to the initial policy. However, we did not observe a single case of such a behavior in our experiments. (ii) It was observed by~\cite{ZikelicLCH22} that methods for jointly learning policies and certificates rely on a good policy initialization. Hence, our repair method is also best suited for cases when the policy is well initialized by some off-the-shelf method (e.g.~with SR at least 90\%).

\section{Conclusion}

In this work, we propose a method for determining the correctness of neural network control policies and certificate functions and for repairing them by utilizing runtime monitoring. Our method applies to the black-box setting and does not assume knowledge of the system dynamics. We present two novel monitoring algorithms, CertPM and PredPM. Our experiments demonstrate the advantage of monitoring policies together with certificate functions and are able to repair neural policies and certificates learned by a state-of-the-art learning-based control method. Interesting directions of future work would be to consider the repair problem for stochastic systems and multi-agent systems. Another interesting direction would be to explore the possibility of deploying predictive monitors towards enhancing the safety of learned controllers upon deployment, i.e.,~at runtime.

\section{Acknowledgments}
This work was supported in part by the ERC project ERC-2020-AdG 101020093.
\bibliography{aaai25}

\newpage
\begin{center}
	\Large\textbf{Appendix}
\end{center}

\section{Runtime Monitoring of Lyapunov Functions}

In this section, we present how to construct $\mathcal{M}_{\textsf{CertPM}}$ in the case of a stability property and a Lyapunov function certificate. Consider a stability property defined by the goal set $\mathcal{X}_g$. Let $\pi$ be a policy and $\mathcal{V}$ be a Lyapunov function. Consider a finite sequence of observed states $x_0, x_1, \dots, x_n$.  $\mathcal{M}_{\textsf{CertPM}}$ checks if all the defining conditions in Proposition~\ref{lc} are satisfied at $x_n$:
\begin{compactitem}
	\item The certificate violation verdict for the Zero upon goal condition in Proposition~\ref{lc} is issued whenever $x_n \in \mathcal{X}_g$ is a goal state but $\mathcal{V}(x_n)  \not=0$. We set $\mathcal{M}_{\textsf{CertPM}}(x_0,x_1,\dots,x_n) = 1$.
	\item The certificate violation verdict for the Strict positivity condition in Proposition~\ref{lc} is issued whenever whenever $\mathcal{V}(x_n) < 0$. We set $\mathcal{M}_{\textsf{CertPM}}(x_0,x_1,\dots,x_n)$.
	\item The certificate violation verdict for the Decreasing condition in Proposition~\ref{lc} is issued as follows. Similarly as for barrier functions, we approximate the Decreasing condition by considering the subsequent observed state $x_{n+1}$ and approximating the Lie derivative via $$\widehat{L_f\mathcal{V}}(x_n)=\frac{\mathcal{V}(x_{n+1}) -\mathcal{V}(x_n)}{t_{n+1} - t_n}.$$ The certificate violation verdict for the Decreasing condition in Proposition~\ref{lc} is issued whenever $x_n\notin\mathcal{X}_g$, $\mathcal{V}(x_n) > 0$ and $\widehat{L_f\mathcal{V}}(x_n) \geq 0$. We set $\mathcal{M}_{\textsf{CertPM}}(x_0,x_1,\dots,x_n)$.
	\item Finally, if none of the three conditions are violated, no violation verdict is issued. We set $\mathcal{M}_{\textsf{CertPM}}(x_0,x_1,\dots,x_n) = 0$.
\end{compactitem}

\section*{Neural Lyapunov Function Repair}

Consider a stability property and suppose that $\mathcal{X}_g$ is a set of goal states in a dynamical system $\Sigma = (\mathcal{X}, \mathcal{U}, f)$. The pseudocode of our monitoring-based repair algorithm is shown in Algorithm~\ref{alg:monitorrepairlyapunov}.

\noindent{\bf Learning-based control with Lyapunov functions.} Before showing how to extract new training data and use it for policy and Lyapunov function repair, we first provide an overview of the general framework for jointly learning neural network policies and Lyapunov functions. Two neural networks are learned simultaneously, by minimizing a loss function which captures each of the defining conditions of Lyapunov functions in Proposition~\ref{lc}. That way, the learning process is guided towards learning a neural control policy that admits a Lyapunov function and hence satisfies the stability property. The loss function contains one loss term for the Zero upon goal and the Decreasing conditions in Proposition~\ref{lc} (Strict positivity condition is enforced by applying a non-negative activation function on the Lyapunov function output):
\begin{equation}\label{eq:losslyapunov}
	\mathcal{L}(\theta, \nu)= \mathcal{L}_{\textrm{G}}(\theta,\nu)+\mathcal{L}_{\textrm{D}}(\theta, \nu),
\end{equation}
where $\theta$ and $\nu$ are vectors of parameters of neural networks $\pi_\theta$ and $\mathcal{V}_\nu$, respectively, and
\begin{equation}\nonumber
	\begin{aligned}
		\mathcal{L}_{\textrm{G}}(\theta,\nu) &= \frac{1}{|D_{\textrm{G}}|}\sum_{x\in D_{\textrm{G}}}\mathcal{V}_\nu(x)^2; \\
		\mathcal{L}_{\textrm{D}}(\theta, \nu) &= \frac{1}{|D_{\textrm{D}}|}\cdot\\ 
		&\sum_{x\in D_{\textrm{D}}}[\max\Big(\widehat{L_{f_\theta}\mathcal{V}}_\nu(x),0\Big) + \max\Big(-\mathcal{L}_\nu(x),0\Big)]. 
	\end{aligned}
\end{equation}
In words, $D_{\textrm{G}}, D_{\textrm{D}} \subseteq \mathcal{X}$ are the training sets of system states used for each loss term and each loss term incurs loss whenever the defining condition in Proposition~\ref{lc} is violated. $D_{\textrm{G}}$ is the data set collected for goal states, and $D_{\textrm{D}}$ is the set for sampled data points. 

\noindent{\bf Monitoring-based neural policy and Lyapunov function repair.} We now show how the verdicts of our monitors are used to obtain new training data that describes certificate violations (recall, for the stability property monitors cannot detect property violations), and how this new training data is used for the policy and Lyapunov function repair. Algorithm~\ref{alg:monitorrepairlyapunov} takes as input the neural network control policy $\pi$ and neural network Lyapunov function $\mathcal{V}$, trained as above. It also takes as input a finite set of time points $t_0 = 0 < t_1 < \dots < t_N$ at which the monitor observes new states, as well as the number $D$ of state trajectories that it monitors. Since we are interested in a stability property, only the CertPM monitor can be used.

Algorithm~\ref{alg:monitorrepairlyapunov} first initializes the monitor $\mathcal{M}$ by constructing the CertPM (line~1), the set of new training data $D_{\textrm{New-data}}$ which is initially empty (line~2) and a set $\tilde{\mathcal{X}}_0 \subseteq \mathcal{X}_0$ of $D$ initial states obtained via sampling from the initial set $\mathcal{X}_0$ (line~3). Then, for each initial state $x_0 \in \tilde{\mathcal{X}}_0$, it executes the dynamical system to obtain a state trajectory $x(t)$ from $x(0) = x_0$ (line~5). For each time point $t_n$, a new state $x_n$ is observed (line~7) and the monitor verdict $\mathcal{M}(x_0,x_1,\dots,x_n)$ is computed (line~8). If $\mathcal{M}(x_0,x_1,\dots,x_n)=1$, i.e. if the monitor issues a certificate violation verdict, the new state $x_n$ is added to the new training dataset $D_{\textrm{New-data}}$ (line~8). Once the new data is collected, it is used to initialize the new training datasets $D_{\textrm{G}}^{\textrm{repair}} = D_{\textrm{New-data}} \cap \mathcal{X}_g$ and $D_{\textrm{N}}^{\textrm{repair}} = D_{\textrm{New-data}} \cap (\mathcal{X}\backslash \mathcal{X}_g)$ (line~9). Finally, the neural network policy $\pi$ and the neural network Lyapunov function $\mathcal{V}$ are repaired by being retraining on the loss function in eq.~\eqref{eq:losslyapunov} with the new training datasets (line~11). If we are only interested in repairing the Lyapunov function $\mathcal{V}$ for a fixed control policy $\pi$, only the network $\mathcal{V}$ is repaired while keeping the parameters of $\pi$ fixed (line~13).

\begin{figure}[t]
	\begin{tabular}{@{}r@{\hspace{.75em}}l@{}}
		\L{}\hspace{-1.25em}\textsc{PolicyRepair} (policy $\pi$, Lyapunov function $\mathcal{V}$, \N
		\L{}\hspace{2em} number $D$, time points $t_0 = 0 < t_1 < \dots < t_N$)\N
		\\[-2.0ex]
		\L{1} $\mathcal{M}\leftarrow \textsc{buildMonitor}(\mathcal{X})$ \scriptsize\textcolor{gray}{$\triangleright$ initializing a CertPM monitor} \N
		\L{2} $D_\textrm{New-data}\leftarrow \emptyset$ \scriptsize\textcolor{gray}{$\triangleright$ new training data collected by the monitor}\N
		\L{3} $\tilde{\mathcal{X}}_0\leftarrow D$ states randomly sampled from $\mathcal{X}_0$ \N
		\L{4}\K{for} $x_0 \in \tilde{\mathcal{X}}_0$ \K{do}\N
		\L{5}\I $x(t)\leftarrow$ state trajectory from $x(0) = x_0$\N
		\L{6}\I\K{for} $n \in \{0,1,\dots,N\}$ \K{do}\N
		\L{7}\I\I $x_n\leftarrow$ observed state at time point $t_n$\N
		\L{8}\I\I \K{if} $\mathcal{M}(x_0,x_1,\dots,x_n) = 1$ \K{then} \N
		\L{9}\I\I\I $D_\textrm{New-data}\leftarrow D_\textrm{New-data} \cup \{x_n\}$\N
		\L{10} $D_{\textrm{G}}^{\textrm{repair}}, D_{\textrm{D}}^{\textrm{repair}} \leftarrow D_{\textrm{New-data}} \cap \mathcal{X}_g, D_{\textrm{New-data}} \cap (\mathcal{X} \backslash \mathcal{X}_g)$\N
		\L{11}\K{if} policy certification and repair \K{then} \scriptsize\textcolor{gray}{$\triangleright$ Problem~1}\N
		\L{12}\I $\pi, \BB \leftarrow$ repair with loss function in eq.~\eqref{eq:losslyapunov} \N
		\L{}\I\I \I and training data $D_{\textrm{G}}^{\textrm{repair}}, D_{\textrm{D}}^{\textrm{repair}}$ \N
		\L{13}\K{else if} certificate certification and repair \K{then} \scriptsize\textcolor{gray}{$\triangleright$ Problem~2}\N
		\L{14}\I $\BB \leftarrow$ repair with loss function in eq.~\eqref{eq:losslyapunov}\N
		\L{} \I\I\I and training data $D_{\textrm{G}}^{\textrm{repair}}, D_{\textrm{D}}^{\textrm{repair}}$
	\end{tabular}
	\captionof{algorithm}{Monitoring-based neural network policy and Lyapunov function repair for stability properties.}
	\label{alg:monitorrepairlyapunov}
\end{figure}

\section*{Additional Experiments}

In addition to the experiments presented in the main part of the paper, we also demontrate the effectiveness of our approach when incoperating Lyapunov functions for stability properties. 
\begin{table}[t]
	\centering
	\caption{Experimental results on repairing the control policy and the Lyapunov function for the drone environment. 
		\label{tb:lyapunov}}
	\begin{tabular}{l|c|c}
		\hline
		& \#$D_\textsc{New-data}$& \textbf{DR}  (\%) \\
		\hline
		Initialized  &-&83.13\\ 
		\hline
		Repair with CertPM &3000&91.32\\ 
		\hline
	\end{tabular}
\end{table}

\begin{figure}\captionsetup[subfigure]{labelformat=empty}
	\subfloat[]{\includegraphics[width = 0.43\textwidth]{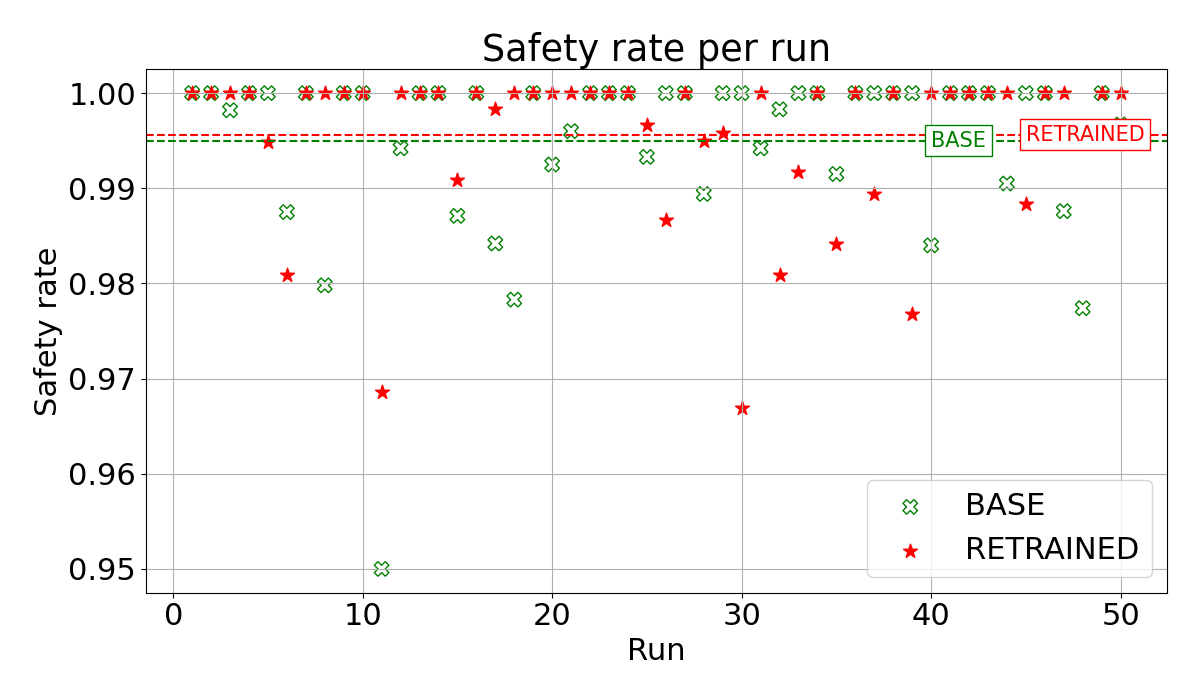}}\vspace{-1.5em}\\
	\subfloat[]{\includegraphics[width = 0.43\textwidth]{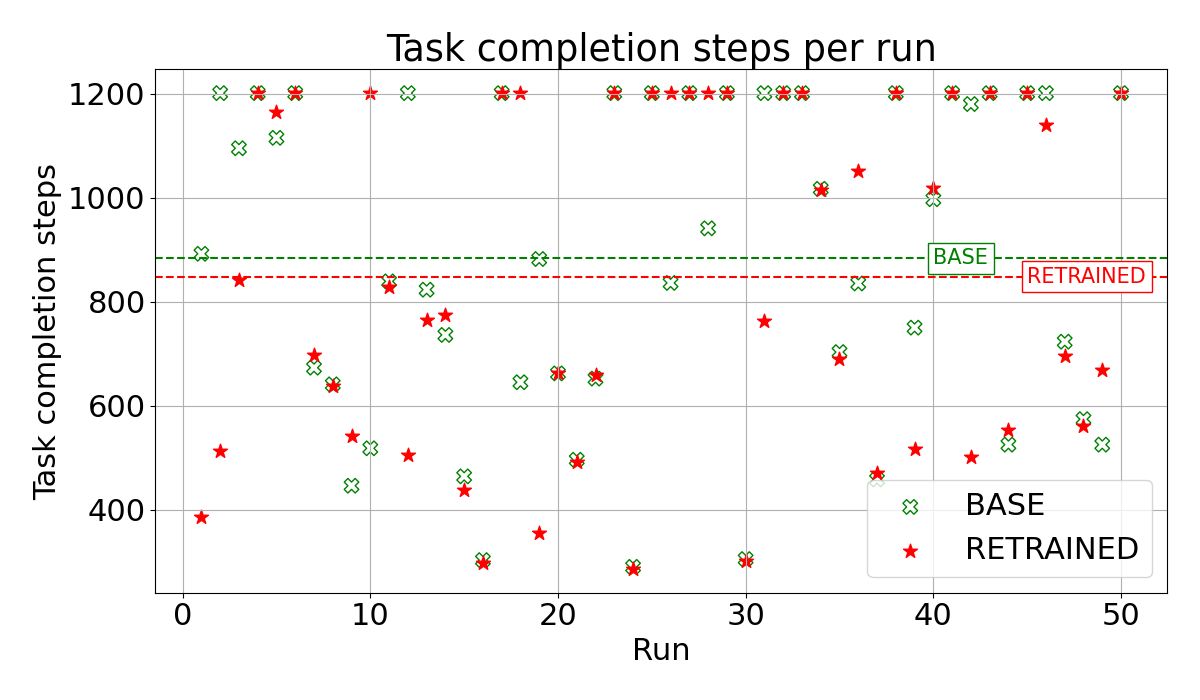}}
	\vspace{-0.5cm}
	\caption{Repairing a reference policy with a Lyapunov function and a barrier function for DroneEnv. The dotted lines represent the average values.\label{fig:refcomp}}
\end{figure}

\noindent{\bf Results: RQ1.} In Table~\ref{tb:lyapunov}, we present the results on repairing a control policy and a Lyapunov function jointly when using CertPM as a monitor. Even though the original codebase did not use a Lyapunov function for stability, we initialized our own by extending the SABLAS framework to also allow learning Lyapunov functions. Table~\ref{tb:lyapunov} shows that our method is able to improve the Decreasing condition rate (DR) of the Lyapunov function from $83.13\%$ to $91.32\%$. For repairing Lyapunov functions, we observe from the experiments that much more training data is required, where a total of $D=2.5\times 10^4$ system executions were monitored, out of which $3000$ violations were detected.
Additionally, in Figure~\ref{fig:refcomp} we present the results of repairing the control policy and the barrier function as trained by~\cite{DBLP:journals/ral/QinSF22} that are publicly available for reference, together with the Lyapunov function learned by our extension of the SABLAS codebase. Again, we use CertPM for the repair. 
We repair the certificate functions with the policy together with $D=10^4$ for each round. As can be seen in Figure~\ref{fig:refcomp}, the repaired policy obtains a higher safety rate over 50 randomly selected runs. Moreover, the repaired control policy reaches the target in fewer steps on average. 

\begin{figure}\captionsetup[subfigure]{labelformat=empty}
	\subfloat[]{\includegraphics[width = 0.43\textwidth]{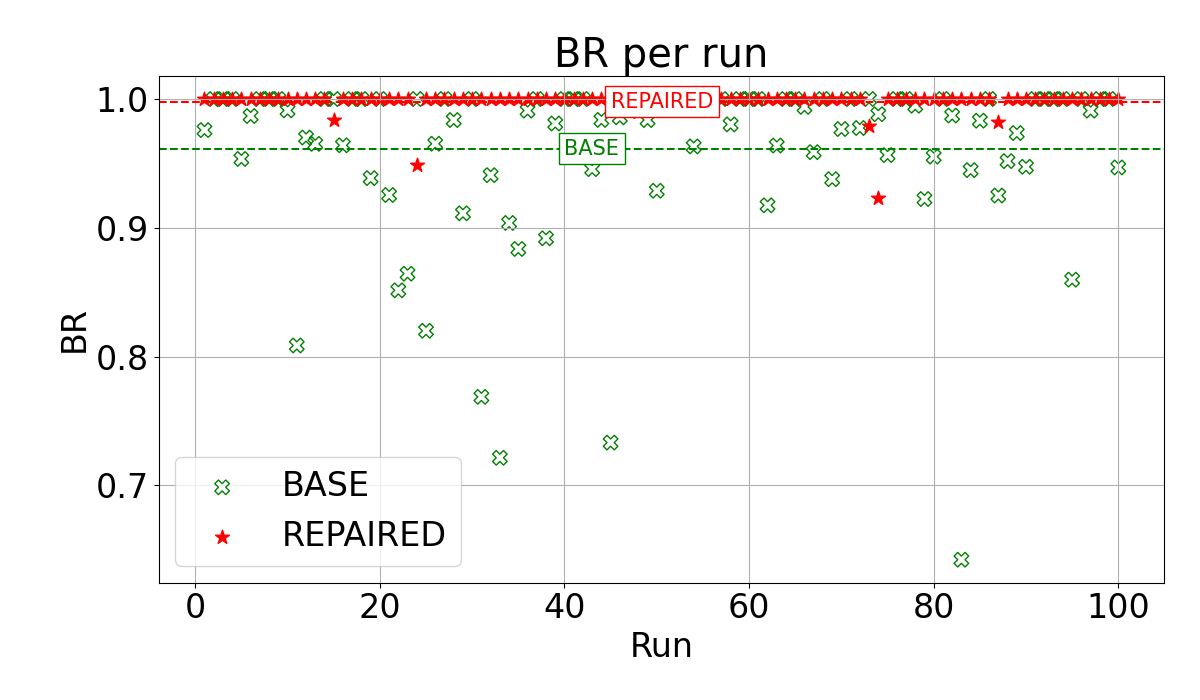}} \vspace{-2em}\\
	\subfloat[]{\hspace{0.35em}\includegraphics[width = 0.43\textwidth]{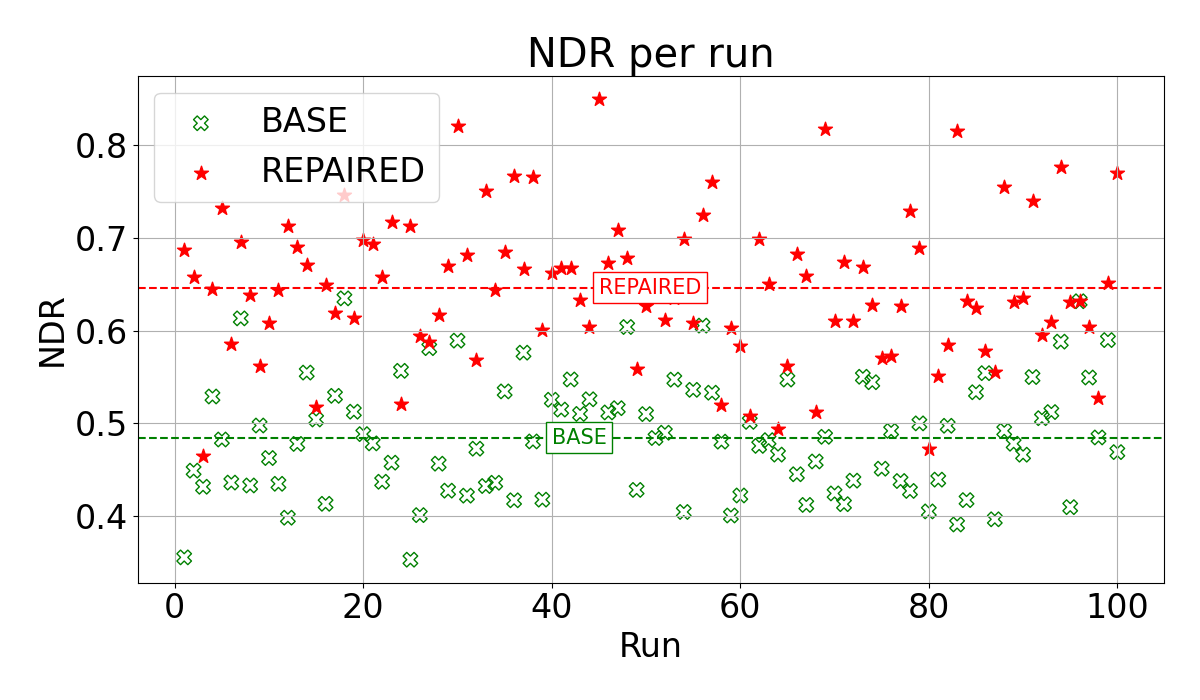}}
	\vspace{-0.5cm}
	\caption{Comparisons of BR and NDR for repairing the barrier function for DroneEnv, training data collected by CertPM. \label{fig:certdrone}}
\end{figure}

\noindent{\bf Results: RQ2.} Figure~\ref{fig:certdrone} illustrates the results of barrier function repair for the drone environment, in the case when the control policy is fixed and we are only interested in repairing the certificate. It shows signiciant improvement in BR and NDR upon the repair of the barrier function.

\begin{figure}
	\includegraphics[width=0.5\textwidth]{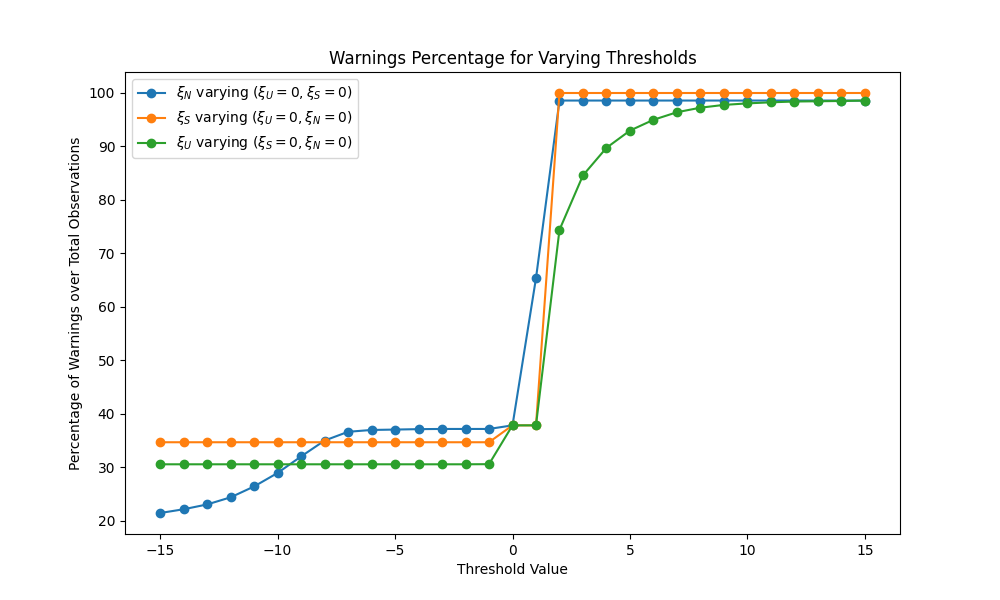}
	\caption{The change in the percentage of warnings over total observations with varying thresholds. For each configuration, we fix two thresholds out of three at zero.\label{fig:thresholds}}
\end{figure}

Additionally, we also conducted experiments with different combinations of varying thresholds $\xi_U, \xi_S,\xi_N$. In each set of experiments, we inspect the proportion of warnings over total states observed by changing the value of one threshold while fixing the values of the other two. This entails how the number of counterexamples collected by our monitor changes, depending on the thresholds chosen by the users. When all three thresholds are set are zero, this is equivalent to using CertPM. For instance, the monitor can be less sensitive towards $v_N$ to allow occasional violations of the non-decreasing condition. This can be achieved by increasing the value of $\xi_N$. 
\end{document}